\documentclass{article}

\usepackage{arxiv}

\usepackage{subcaption}
\usepackage{amsmath}

\usepackage[utf8]{inputenc} % allow utf-8 input
\usepackage[T1]{fontenc}    % use 8-bit T1 fonts
\usepackage{hyperref}       % hyperlinks
\usepackage{url}            % simple URL typesetting
\usepackage{booktabs}       % professional-quality tables
\usepackage{amsfonts}       % blackboard math symbols
\usepackage{nicefrac}       % compact symbols for 1/2, etc.
\usepackage{microtype}      % microtypography
\usepackage{cleveref}       % smart cross-referencing
\usepackage{graphicx}
\usepackage{natbib}
\usepackage{doi}

\title{WoodYOLO: A Novel Object Detector for Wood Species Detection in Microscopic Images}

\author{ \href{https://orcid.org/0000-0002-7523-5694}{\includegraphics[scale=0.06]{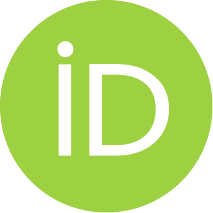}\hspace{1mm}Lars~Nieradzik}\\
	Image Processing Department\\
	Fraunhofer ITWM\\
	Fraunhofer Platz 1, 67663, Kaiserslautern\\
	\texttt{lars.nieradzik@itwm.fraunhofer.de}\\
	\And
	\href{https://orcid.org/0000-0002-9821-1636}{\includegraphics[scale=0.06]{orcid.pdf}\hspace{1mm}Henrike~Stephani} \\
	Image Processing Department\\
	Fraunhofer ITWM\\
	Fraunhofer Platz 1, 67663, Kaiserslautern\\
	\texttt{henrike.stephani@itwm.fraunhofer.de}\\
	\And
	\href{https://orcid.org/0009-0001-7547-269X}{\includegraphics[scale=0.06]{orcid.pdf}\hspace{1mm}Jördis ~Sieburg-Rockel} \\
	Thünen Institute of Wood Research\\
	Leuschnerstraße 91, 21031, Hamburg\\
	\texttt{joerdis.sieburg-rockel@thuenen.de} \\
	\And
	\href{https://orcid.org/0009-0009-6611-3140}{\includegraphics[scale=0.06]{orcid.pdf}\hspace{1mm}Stephanie~Helmling} \\
	Thünen Institute of Wood Research\\
	Leuschnerstraße 91, 21031, Hamburg\\
	\texttt{stephanie.helmling@thuenen.de} \\
	\And
	\href{https://orcid.org/0009-0007-2249-2797}{\includegraphics[scale=0.06]{orcid.pdf}\hspace{1mm}Andrea~Olbrich} \\
	Thünen Institute of Wood Research\\
	Leuschnerstraße 91, 21031, Hamburg\\
	\texttt{andrea.olbrich@thuenen.de} \\
	\And
	\href{https://orcid.org/0009-0000-0112-6113}{\includegraphics[scale=0.06]{orcid.pdf}\hspace{1mm}Stephanie~Wrage} \\
	Thünen Institute of Wood Research\\
	Leuschnerstraße 91, 21031, Hamburg\\
	\texttt{stephanie.wrage@thuenen.de} \\
	\And
	\href{https://orcid.org/0000-0002-1327-1243}{\includegraphics[scale=0.06]{orcid.pdf}\hspace{1mm}Janis~Keuper} \\
	Institute of Machine Learning and Analysis (IMLA)\\
	Offenburg University\\
	Badstr. 24, 77652, Offenburg\\
	\texttt{jkeuper@ad.hs-offenburg.de}\\
}

\renewcommand{\shorttitle}{Challenging the Black Box}

\hypersetup{
pdftitle={A template for the arxiv style},
pdfsubject={q-bio.NC, q-bio.QM},
pdfauthor={David S.~Hippocampus, Elias D.~Striatum},
pdfkeywords={First keyword, Second keyword, More},
}

\begin{document}
\maketitle

\begin{abstract}Wood species identification plays a crucial role in various industries, from ensuring the legality of timber products to advancing ecological conservation efforts. This paper introduces WoodYOLO, a novel object detection algorithm specifically designed for microscopic wood fiber analysis. Our approach adapts the YOLO architecture to address the challenges posed by large, high-resolution microscopy images and the need for high recall in localization of the cell type of interest (vessel elements).
Our results show that WoodYOLO significantly outperforms state-of-the-art models, achieving performance gains of 12.9\% and 6.5\% in F2 score over YOLOv10 and YOLOv7, respectively. This improvement in automated wood cell type localization capabilities contributes to enhancing regulatory compliance, supporting sustainable forestry practices, and promoting biodiversity conservation efforts globally.
\end{abstract}

% keywords can be removed

%%%%%%%%% BODY TEXT

\section{Introduction}

Global deforestation is a cause of biodiversity loss and climate change. The European Union's recently adopted EU Deforestation Regulation (EUDR, \cite{european2023regulation}), which replaces the EU Timber Regulation (EUTR), requires that products traded in the EU are based on deforestation-free supply chains. This increases the demand for confirming the declaration of wood products regarding the wood species and origin. 

\begin{figure}[htb]
    \centering
    \includegraphics[width=\textwidth]{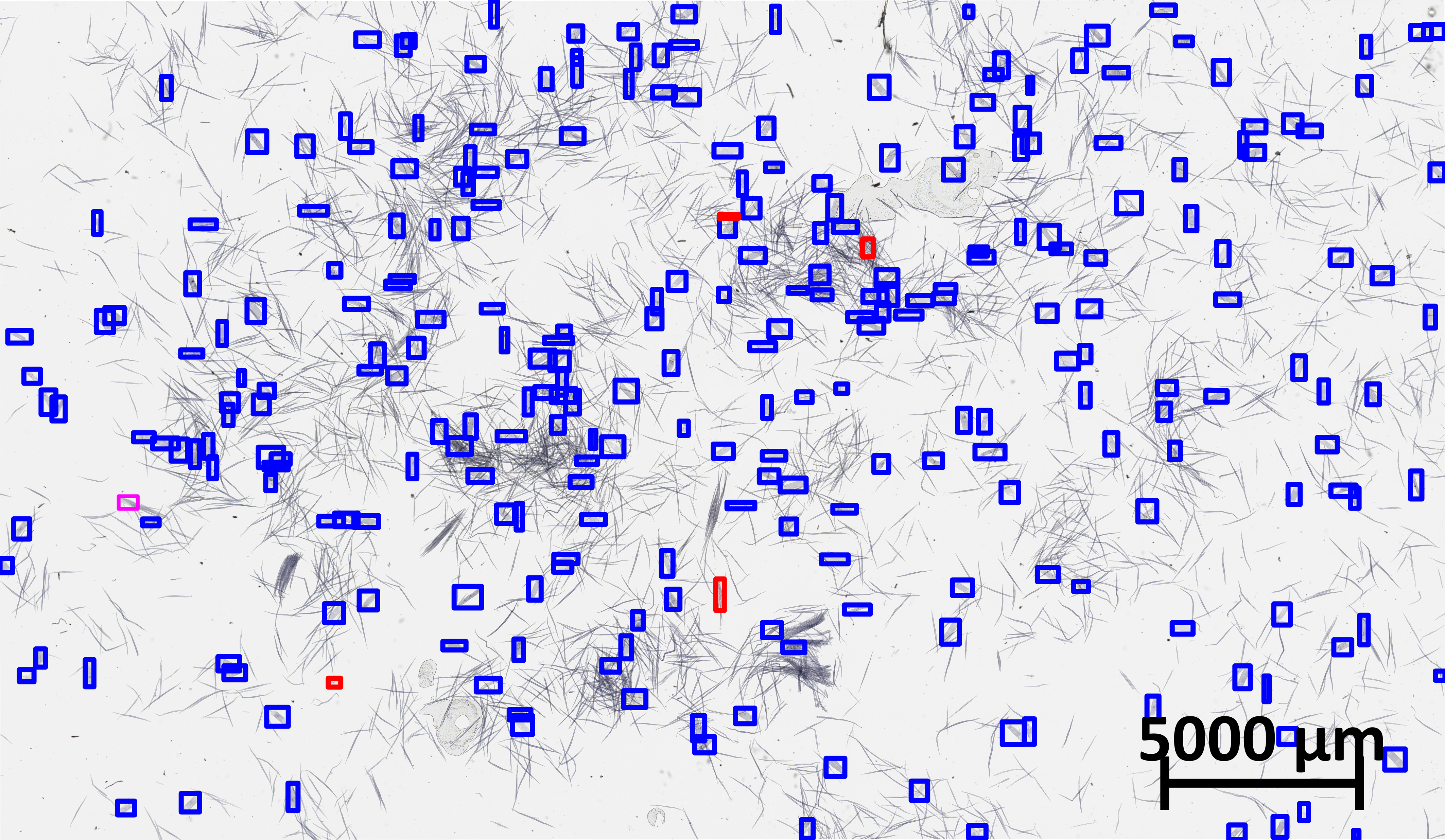}
    \caption{Microscope image of macerated hardwood cells including vessel elements. Blue boxes indicate the correctly localized vessel locations by WoodYOLO, the light purple box indicate one false negative and red boxes denote false positives that were not annotated by wood anatomists. WoodYOLO significantly speeds up the manual annotation process by automatically identifying hundreds of vessel elements.}
    \label{fig:wood_vessels}
\end{figure}

This is a particular challenge for paper products where the DNA is destroyed and different pulps are mixed during production. Therefore, neither genetics, stable isotopes nor NIR spectroscopy can be used. It is therefore not possible to analyze the origin \citep{tsuchikawa2015review, schmitz2020overview}. Recently a new, very complex chemotaxonomic method for determining wood species was introduced for the first time \citep{FlaigBergerWenigOlbrichSaake+2023+860+878}. But the standard analysis for checking the declared wood species in paper is still the anatomy \citep{helmling2018atlas, ilvessalo1995fiber}. After sample preparation, the microscopic examination of the cell characteristics by experts is time-consuming and requires a high level of personal experience. \Cref{fig:wood_vessels} shows an example of a microscope image of macerated cells that is analyzed. The limited number of experts in the field makes it challenging to meet the increasing demand for wood species verification \citep{ruffinatto2019atlas}.

To address these challenges, recent advancements in computer vision and machine learning offer promising avenues for automating wood species identification. Machine learning techniques, particularly deep neural networks, have shown remarkable capabilities in analyzing large-scale image datasets and extracting intricate features crucial for species classification \citep{silva2022computer}. However, while automated systems exist for macroscopic wood analysis \citep{MyWood, ravindran2020xylotron, wiedenhoeft2020xylophone}, automated methods for microscopic analysis of fibrous materials like paper are still nascent \citep{nieradzik2023automating}.

A deep learning-based approach specifically targeting the detection and classification of vessel elements in microscopic images of macerated wood samples was recently presented \citep{nieradzik2023automating}. These efforts have highlighted the potential of automation to streamline what has traditionally been a manual task. However, existing methods often face challenges such as suboptimal recall and high demands on computational power, especially when processing large and high-resolution microscopic images.

To address these limitations and further advance automated wood species identification, we present WoodYOLO, a novel object detection algorithm specifically designed for microscopic wood fiber analysis. WoodYOLO builds upon the YOLO (You Only Look Once) architecture, incorporating tailored optimizations to enhance performance in high-resolution microscopy.

Our algorithm introduces several key innovations:

\begin{itemize}
    \item Customized YOLO-based architecture specifically optimized for microscopic images, achieving significant performance gains over YOLOv10 and YOLOv7 by 12.9\% and 6.5\% respectively in terms of F2 score, while using around 3-4x less VRAM.
    \item Introduction of a novel anchor box specification method, where users define only the maximum width and height of objects. This approach improves F2 score by 0.7\%.
    \item Comprehensive evaluation of various architectural decisions in modern object detectors. Our findings reveal that optimizations designed for general datasets like COCO \citep{lin2015microsoftcococommonobjects} may not always translate to improved performance in real-world datasets or different domains.
\end{itemize}

By advancing automated wood species identification capabilities, our work contributes to enhancing regulatory compliance, supporting sustainable forestry practices, and promoting biodiversity conservation efforts globally. WoodYOLO represents a significant step towards developing scalable, reliable and efficient methods for wood species identification in microscopic images of fibrous materials.

\section{Related Work}

The automated identification of wood species in microscopic images of fibrous materials has gained significant attention in recent years. This interest is driven by the need for efficient and accurate methods to support global wood fiber product controls.

A pioneering approach for the identification of hardwood species in microscopic images using deep learning techniques was introduced by \cite{nieradzik2023automating}. They developed a methodology for generating a large dataset of macerated wood references, focusing on nine hardwood genera. This approach utilized a two-step process: first, detecting vessel elements using YOLOv7 \citep{wang2022yolov7trainablebagoffreebiessets}, and then classifying these elements using convolutional neural networks (CNNs).

While the localization of objects achieved promising results, there remains room for improvement. Recently developed object detection algorithms, particularly those based on transformers, such as the DETR (DEtection TRansformer) model family \citep{carion2020endtoendobjectdetectiontransformers, zhao2024detrsbeatyolosrealtime, zhang2022dinodetrimproveddenoising, ouyangzhang2022nmsstrikes}, have shown potential. However, they have not seen widespread use due to higher time complexity, slower training speeds or lower mAP on real-world datasets.

Another line of research is the continuation of YOLO. It is important to note that a higher version number in YOLO does not necessarily indicate an improvement; instead, different techniques are applied, which may or may not work on particular datasets. Since the original YOLO publication \citep{redmon2016lookonceunifiedrealtime}, only YOLOv2 \citep{redmon2016yolo9000betterfasterstronger} and YOLOv3 \citep{redmon2018yolov3incrementalimprovement} were developed by the original authors. Other versions have been introduced by different institutes or companies, including YOLOv4 \citep{bochkovskiy2020yolov4optimalspeedaccuracy}, Scaled-YOLOv4 \citep{wang2021scaledyolov4scalingcrossstage}, YOLOX \citep{ge2021yoloxexceedingyoloseries}, YOLOv6 \citep{li2022yolov6singlestageobjectdetection}, DAMO-YOLO \citep{xu2023damoyoloreportrealtime}, YOLOv9 \citep{wang2024yolov9learningwantlearn}, YOLOv10 \citep{wang2024yolov10realtimeendtoendobject}, PP-YOLO \citep{long2020ppyoloeffectiveefficientimplementation}, PP-YOLOv2 \citep{huang2021ppyolov2practicalobjectdetector}, and PP-YOLOE \citep{xu2022ppyoloeevolvedversionyolo}. Notably, YOLOv5 and YOLOv8 have never been published. In our method section, we will analyze some of the different components found in these papers.

A recent study by \cite{Qamar2024} titled "Segmentation and characterization of macerated fibers and vessels using deep learning" demonstrated the application of YOLOv8 for analyzing microscopy images of wood fibers.

In most practical machine learning research and data competitions, YOLO remains the state-of-the-art. Therefore, our focus is on developing an object detector based on this literature. Our current work builds upon these foundations by introducing a novel object detection algorithm specifically tailored for vessel element detection in microscopic images of fibrous materials. By designing our detection algorithm with this task in mind, we can make better optimizations and avoid focusing on general-purpose detection datasets such as COCO.

Although there are numerous papers in the microscopy and satellite imaging literature that adapt YOLO for high-resolution image analysis, they generally rely on the original YOLO code base and make only minor changes. For example, \cite{LpezFlrez2023, Aldughayfiq2023} adapted YOLOv5 for cell counting. There are also various studies in the field of satellite images in which YOLO \citep{rs15153770, 9534343} has been slightly modified. As a result, the improvements compared to the baseline are often only marginal. In contrast, we have developed our version of YOLO from scratch and tested components from different versions. This allows for more significant and customized improvements specifically for our application.

\section{Materials and methods}

Frequently processed woods that are cultivated in plantations for pulp, paper and fiberboard production were selected, such as poplar or eucalypt. The exact genera can be found in \cite{nieradzik2023automating}. Vouchered specimens of the Thünen Institute's wood collection and other documented sources served as reference material for training and testing. 
Analogous to pulp production, the cell structure of the wood tissue was broken down into individual cells by maceration according to the method of \cite{franklin1945preparation}. At least 3 macerates per genus were produced. Maceration and staining are described in detail in \cite{helmling2016qualitative} and \cite{helmling2018atlas}. For each macerate, 20 slides were prepared. Ten of these were stained with Alexander Herzberg solution and ten with nigrosine (1 wt\%).

Our detection framework is tailored to localize vessel elements in microscopic images, a crucial step for automating hardwood species identification in fibrous materials. Vessel elements, the cell elements for conducting water in deciduous trees, contain characteristic morphological features that differ within the genera, in contrast to fibers. We adapted the YOLO architecture for this domain, addressing the challenges posed by large image sizes (up to 54,000 x 31,000 pixels) and the need for high recall. Unlike algorithms such as DETR, which do not scale well for very large images and have slower training times, YOLO has proven effective in real-world applications, making it a suitable choice for our task.

Although the YOLO family includes various models optimized for general datasets like COCO, these models are not directly applicable to our problem due to their design for multiple classes and general-purpose images. Therefore, we customized YOLO by integrating components from different versions to optimize it for vessel detection without the need for classification.

In this section, we describe our model's architecture, loss function, metric, and additional approaches evaluated to enhance detection performance.

\subsection{Architecture}

\begin{figure}[ht]
  \centering
  \includegraphics[scale=0.315]{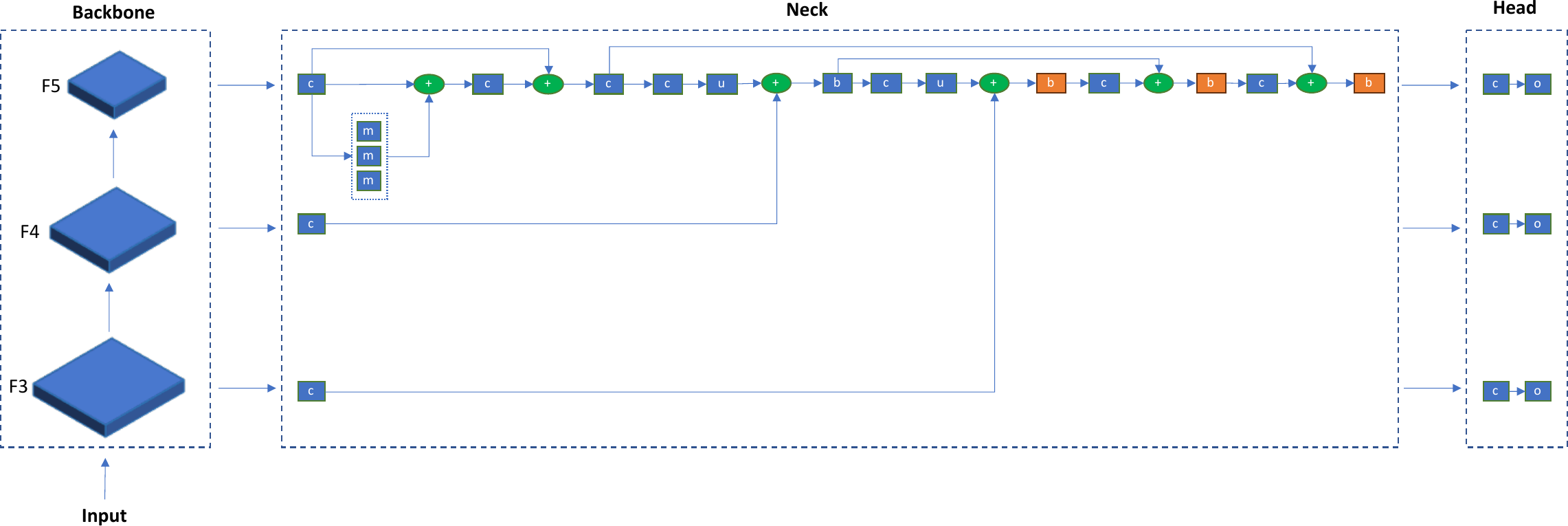}
  \caption{Detection architecture based on YOLOv7-tiny \citep{wang2022yolov7trainablebagoffreebiessets}. "c" = Convolution with BN and ReLU, "+" = Concatenation, "m" = MaxPooling, "u" = Upsampling, "b" = Concatenation Block, "o" = single convolution with 5 outputs (x, y, width, height, confidence). Orange denotes an output in the neck of the model, which is given to the head. There are in total three outputs.}
  \label{fig:detection_pipeline}
\end{figure}

Our model architecture begins with selecting a backbone capable of efficiently extracting features from large microscopic images. The backbone processes the input to generate multi-scale feature maps. We tested several backbones such as VGG11 \citep{simonyan2015deepconvolutionalnetworkslargescale}, ConvNext \citep{liu2022convnet2020s}, and ResNet \citep{he2015deepresiduallearningimage}, and combined their feature maps through a component known as the neck, which outputs three feature maps. Although more than three feature maps can be used, our evaluation showed no significant advantage in doing so.

Our neck architecture is based on YOLOv7-tiny. We also tested YOLOX's CSPNet\citep{wang2019cspnetnewbackboneenhance} but found the former to be better. The use of a smaller architecture is due to the need for memory efficiency. Since we want to train the network with a higher image resolution than the usual 640x640 or 1280x1280, we need to reduce the memory requirements. Also, deeper networks are usually chosen when many features are needed to distinguish between different classes. Here it is only a matter of finding objects without the need for classification. Therefore, simpler networks work better.

\begin{figure}[ht]
    \centering
    \includegraphics[width=0.3\linewidth]{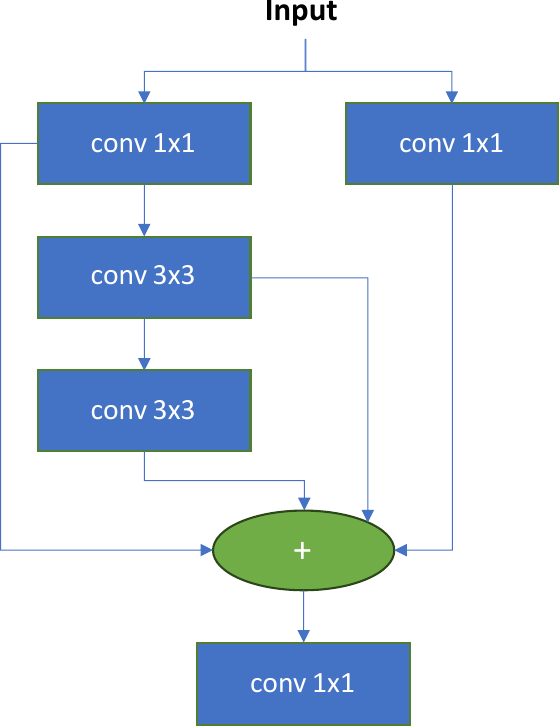}
    \caption{The "b" concatenation block consists of convolutions of kernel size 3x3 and 1x1. Each convolution is followed by batch normalization and ReLU activation. The "+" means concatenation.}
    \label{fig:parallelconv}
\end{figure}

\Cref{fig:detection_pipeline} shows that our neck consists of several convolutional layers that are combined in different ways. A "c" block consists of a simple convolution followed by a batch normalization and a ReLU function. The "b" block consists of parallel convolutions that are combined by concatenation. \Cref{fig:parallelconv} shows the "b" block in detail.

The three orange blocks in \cref{fig:detection_pipeline} indicate the outputs of the neck. These three blocks are then used as inputs for the head. The outputs have different dimensions as a higher stride size is used for some of the convolutions.

The head produces the predictions of the neural network. It consists of only one convolutional block and one output convolution. A decoupled head, such as the one used in YOLOX, has not proven to be better in our case.

For each feature map, the head produces an output tensor $f_i$ of dimensions $g_{h_i} \cdot g_{w_i} \times 5$, where $g_{h_i}$ and $g_{w_i}$ denote the grid height and width for the $i$-th layer. Each grid cell in $f_i$ predicts five parameters: the center x-coordinate, center y-coordinate, width, height, and object confidence. These outputs are transformed as follows:

\[
\begin{aligned}
    x_c &= 2\sigma(f_{i,\cdot,1}) - 0.5, \\
    y_c &= 2\sigma(f_{i,\cdot,2}) - 0.5, \\
    w &= \sigma(f_{i,\cdot,3})^2 \cdot g_{w_i} \cdot m_w, \\
    h &= \sigma(f_{i,\cdot,4})^2 \cdot g_{h_i} \cdot m_h, \\
    o &= \sigma(f_{i,\cdot,5}),
\end{aligned}
\]

where $m_w$ and $m_h \in [0, 1]$ are hyperparameters defining the maximum width and height of the object. For instance, $m_w=0.1$ means an object can be at most 10\% of the total image width. This is similar to having a single anchor box of a maximum specific size.

The advantage of using two hyperparameters instead of anchor boxes is that no techniques such as clustering \citep{redmon2016yolo9000betterfasterstronger} have to be used to determine them. In addition, the loss function is much simpler and the training speed is higher.

The sigmoid function $\sigma(\cdot)$ ensures $x_c$ and $y_c$ are offsets within the grid, while $w$ and $h$ define bounding box dimensions. The confidence score $o$ indicates the likelihood of a bounding box's presence at each location.

$x_c$ and $y_c$ are scaled here between $[-0.5, 1.5]$. This allows the model to shift the center of the box half to the left or right.

In the prediction phase, $x_c$ and $y_c$ offsets are adjusted by adding the grid indices $\{0, 1, \dots, g_w\}$ and $\{0, 1, \dots, g_h\}$, respectively. Coordinates are scaled to the original image size by multiplying $x_c$, $y_c$, $w$, and $h$ by $\frac{s_h}{g_{h_i}}$ and $\frac{s_w}{g_{w_i}}$, where $s_h$ and $s_w$ are the input image dimensions.

\subsection{Loss Function}

Our loss function consists of two components:

\[
L = L_r + L_p\,,
\]

where $L_r$ is the regression loss and $L_p$ is the classification loss.

\paragraph{Regression Loss}
The regression loss measures the alignment between predicted bounding boxes $\hat{b}$ and ground truth $b$ using the Intersection over Union (IoU):

\[
L_r = \frac{1}{m}\sum_{i=1}^n\sum_{j=1}^m (1 - \operatorname{IoU}(\hat{b}_{i,j}, b_{i,j})),
\]

where $n$ is the number of feature pyramid layers (in our case, $n = 3$) and $m$ is the number of bounding boxes. The regression loss is either evaluated with the corresponding bounding box at that grid cell or additionally with neighboring grid cells (multi-positives).

There are different variants of IoU: Complete IoU (cIoU)\citep{zheng2021enhancinggeometricfactorsmodel}, Distance IoU (DIoU)\citep{zheng2019distanceioulossfasterbetter}, Generalized IoU (GIoU) \citep{rezatofighi2019generalizedintersectionunionmetric} and standard IoU. In the evaluation section, we evaluate the different approaches to see which maximizes our metric.

\paragraph{Classification Loss}
The classification loss evaluates the confidence score $\hat{o}$ using binary cross entropy (BCE), with the ground truth confidence $o$ derived from IoU:

\[
L_p = \frac{1}{m}\sum_{i=1}^n\sum_{j=1}^m \operatorname{BCE}(\hat{o}_{i,j}, \operatorname{IoU}(\hat{b}_{i,j}, b_{i,j})).
\]

Unlike the regression loss, we evaluate BCE at all locations of the grid. However, we set $\operatorname{IoU}(\hat{b}_{i,j}, b_{i,j}) = 0$ when there is no ground truth box at a specific grid cell. 

\subsection{Metric}

The predominant metric in object detection is average precision (AP) \citep{10.1007/s11263-009-0275-4} computed at different thresholds, which summarizes both precision and recall:

\[
\text{AP} = \int_0^1 p(r) \, dr\,,
\]

where $r$ denotes recall and $p(r)$ denotes precision as a function of recall. A detection is considered correct if the IoU between the predicted and the true bounding box exceeds a predefined threshold.

In our specific application, however, the use of AP would not be a good choice. Recall takes precedence over precision as our goal is to find all objects.

\begin{figure}[htbp]
    \centering
        \includegraphics[width=\textwidth]{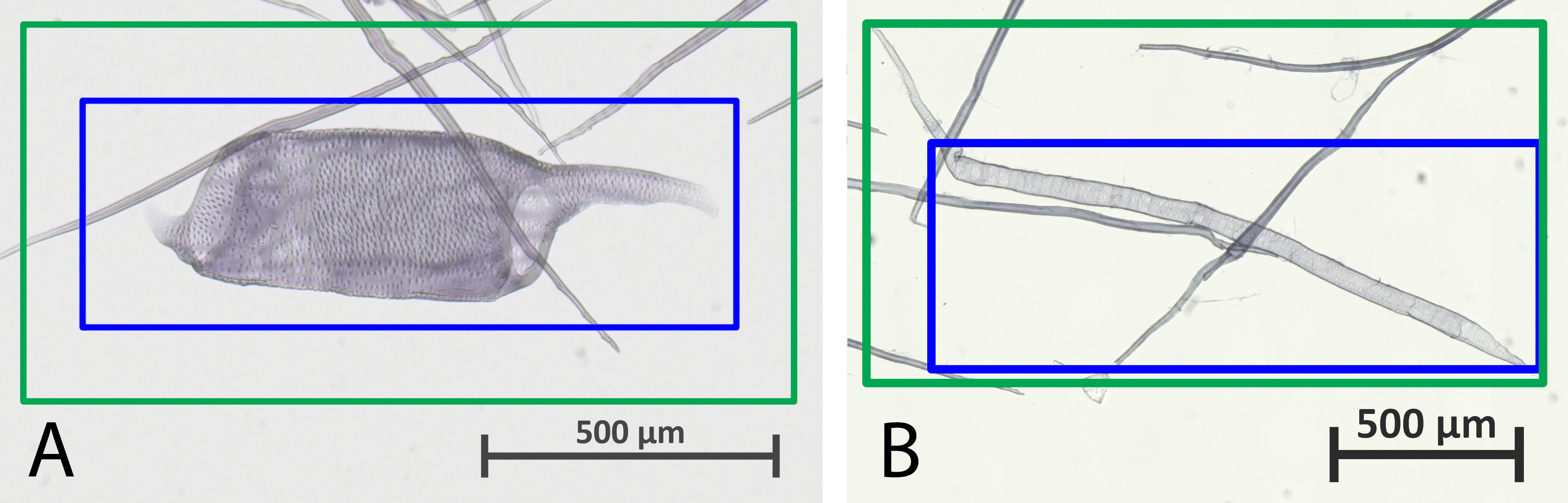}
    \caption{Comparison of predicted bounding boxes (blue) and ground truth boxes (green). A high IoU threshold can result in both predicted boxes being rated as errors. \textbf{(A)} The overlap is below 0.5. Due to incorrect annotations, the predicted bounding boxes are sometimes more accurate. \textbf{(B)} Imperfect prediction as the end of the object (vessel element) is not detected.}
    \label{fig:iouok}
\end{figure}

Furthermore, we are less interested in an exact overlap with the ground truth. Minor shifts or size variations in the bounding box should not be penalized by the metric. Therefore, we want to consider only a single low IoU threshold. Often AP is computed at multiple thresholds.

Hence, we propose an alternative metric: the F2 score, which is computed with a fixed IoU threshold of 0.3. This choice emphasizes recall over precision. False positives can be handled in a postprocessing step by training a classifier to distinguish between correct and wrong detections. We see in \cref{fig:iouok} two examples where the overlap of 30\% is sufficient.

While the usual threshold is 0.5, we choose a lower threshold of 0.3. This threshold takes into account the fact that perfect alignment with the ground truth bounding box is not essential for our objectives.

\subsection{Additional Approaches}

We explored several innovations from the YOLO series to further enhance our detection framework, evaluating their impact on performance. Some of these results will be shown in the evaluation section.

\paragraph{Center Sampling and Multi-Positives}

We explored the use of neighboring grid cells for matching ground truth boxes, a technique known in the literature as multi-positives \citep{ge2021yoloxexceedingyoloseries} or center sampling \citep{tian2019fcosfullyconvolutionalonestage}.

In the standard loss function $L_r$, we compute the IoU loss only between boxes at coordinates \((i,j)\). Center sampling extends this concept by also comparing boxes at \((i+k_1, j+k_2)\), where \( k_1 \) and \( k_2 \) are integer offsets. The ground truth box is duplicated for these new coordinates \((i+k_1, j+k_2)\) to make a comparison with the ground truth box at those positions possible. We investigated three variants:

\[
\text{0 Neighbors:}
\begin{bmatrix}
0 & 0 & 0 \\
0 & \circ & 0 \\
0 & 0 & 0 \\
\end{bmatrix}
\quad\quad
\text{2 Neighbors:}
\begin{bmatrix}
0 & \times & 0 \\
0 & \circ & \times \\
0 & 0 & 0 \\
\end{bmatrix}
\quad\quad
\text{4 Neighbors:}
\begin{bmatrix}
0 & \times & 0 \\
\times & \circ & \times \\
0 & \times & 0 \\
\end{bmatrix}
\]

Here, \(\circ\) denotes the original bounding box, while \(\times\) represents neighboring boxes and $0$ means "empty cell". For the 0 neighbors configuration, the loss \( L_r \) remains unchanged as it only considers the original box. In the 2 neighbors configuration, the nearest bounding boxes within the grid are selected, in this case, the right and upper boxes. For the 4 neighbors configuration, we use bounding boxes from all directions: left, right, up, and down. Note that the diagonal boxes are never selected.

Since object detection is a one-to-many mapping (one ground-truth box corresponds to many correctly predicted boxes), this strategy attempts to simulate this mapping using the loss function.

\paragraph{Label Assignment}

Bounding boxes are predicted for every feature map. The use of center sampling further increases the number of predicted boxes. To manage this increase of bounding boxes, we evaluated label assignment strategies designed to reduce the number of valid boxes per object.

We experimented with modern label assignment techniques such as SimOTA and TAL \citep{ge2021otaoptimaltransportassignment, feng2021toodtaskalignedonestageobject}. However, these methods did not yield improved results in our scenario. We attribute this to our metric, which prioritizes maximizing recall rather than balancing precision and recall.

\paragraph{Auxiliary Head Loss}

Deep supervision techniques, such as those used in YOLOv7 \citep{wang2022yolov7trainablebagoffreebiessets}, involve adding auxiliary losses to guide deeper networks. Our experiments with additional model layers showed no benefit, so this approach was excluded from our final model.

\paragraph{Anchor Boxes}

Anchor boxes, introduced in YOLOv2 \citep{redmon2016yolo9000betterfasterstronger}, are used to predict object locations. Consistent with YOLOX findings \citep{ge2021yoloxexceedingyoloseries}, our results showed no improvement with anchor boxes, leading us to exclude them for simplicity. Instead, we incorporate parameters \( m_h \) and \( m_w \) in the range \([0, 1]\) to constrain the predicted width and height of bounding boxes, as discussed previously.

\paragraph{NMS-Free Detection}

NMS-free approaches from models like YOLOv10 did not perform as well in our tests. We retained traditional Non-Maximum Suppression (NMS) for its robustness and simplicity.

\paragraph{Training Strategies}

Techniques such as mosaic augmentation and gradient accumulation, which are effective in other YOLO implementations, did not significantly improve detection in our application. Therefore, they were excluded from the final model configuration.

\section{Results}

We evaluate WoodYOLO on a dataset constructed for automating the detection and identification of vessel elements in hardwood species, a critical step toward wood species classification. Vessel elements are the water-conducting cells in hardwoods, that differ from genus to genus due to their characteristic morphological features. These vessel elements provide vital information for wood identification  and are easily to distinguish from other cell types like fibers or parenchyma cells.

In this paper, we are specifically concerned with improving the localization of these vessel elements. The dataset comprises high-resolution microscope images of macerated hardwood samples, captured with a ZEISS Axioscan 7 microscope. Each image, originally in the czi format with a resolution of approximately 54,000 x 31,000 pixels and file size of 1 GB, was scaled down by 10\% (5,400 x 3,100 pixels) to enhance training efficiency and reduce memory usage. The final dataset consists of 767 images annotated with 118,287 bounding boxes identifying vessel elements.

Only the third of five focal planes of each image was utilized for training, as additional planes did not contribute significant information for detecting the vessel elements. The annotated dataset was split into 613 images for training and 154 images for validation. We have conducted initial experiments with 5-fold cross-validations, but found that the metrics are relatively stable across different folds. Due to time constraints, we use a simple train-validation split.

In this section, we evaluate the performance of our vessel detection framework across various configurations and compare it to other state-of-the-art models. The evaluations were conducted using the F2 score at a fixed IoU threshold of 0.3, as described before.

\subsection{Detection Model and backbone comparison}

Since we use YOLO as a basis, it is useful to compare our model with other YOLO variants. In \cref{tab:architecture-comparison}, we present the F2 scores for different detection models.

\begin{table}[ht]
  \centering
  \begin{tabular}{lc}
    \toprule
    Architecture & F2 Score \\
    \midrule
    YOLOv10-S & 0.691 \\
    YOLOv10-M & 0.719 \\
    YOLOv7-W6 & 0.783 \\
    YOLOv7-tiny & 0.723 \\
    Ours & \textbf{0.848} \\
    \bottomrule
  \end{tabular}
  \caption{Comparison of detection models based on the F2 score. "Ours" refers to our best WoodYOLO configuration.}
  \label{tab:architecture-comparison}
\end{table}

\begin{table}[ht]
  \centering
  \begin{tabular}{lcc}
    \toprule
    Backbone & Params (M) & F2 Score \\
    \midrule
    YOLOv7-tiny \citep{wang2022yolov7trainablebagoffreebiessets} & 9.77 & 0.8146 \\
    VGG11-bn \citep{simonyan2015deepconvolutionalnetworkslargescale} & 16.59 & \textbf{0.8316} \\
    RepVGG-A0 \citep{ding2021repvggmakingvggstyleconvnets} & 15.52 & 0.8168 \\
    ResNet-18 \citep{he2015deepresiduallearningimage} & 18.48 & 0.8096 \\
    EfficientNet-B0 \citep{tan2020efficientnetrethinkingmodelscaling} & 10.78 & 0.8198 \\
    ConvNeXt-Nano \citep{liu2022convnet2020s} & 22.33 & 0.8284 \\
    \bottomrule
  \end{tabular}
  \caption{Comparison of different backbone networks. We used 2 neighbors for this experiment.}
  \label{tab:backbone-comparison}
\end{table}

 Our customized YOLO variant outperforms other models, achieving an F2 score of 0.848, highlighting its superior ability to detect vessel elements in large microscopic images.

 The parameters of YOLOv10 and YOLOv7 have both been optimized. It is worth noting that we use a resolution of 5184x5184 pixels for the second-best model YOLOv7-W6, which requires the use of an A100. Our model uses 2048x2048 and can be trained with less than 10 GB of VRAM.

We also evaluated various backbone networks to determine their impact on detection performance. Table \ref{tab:backbone-comparison} summarizes the results, while including the number of parameters.

The VGG11-bn backbone yielded the highest F2 score (0.8316) while maintaining a reasonable parameter count and VRAM usage. All the other backbones except YOLOv7-tiny have much higher VRAM requirements as they use skip connections, more complex activation functions or special layers like Squeeze-and-Excitation blocks \citep{hu2019squeezeandexcitationnetworks}. The simplicity of VGG makes it possible to scale it easier to higher resolutions.

\subsection{Effect of Neighboring Cells and IoU Loss function}

We assessed the impact of considering neighboring grid cells (multi-positives) for matching ground truth boxes. As shown in Table \ref{tab:neighboring-cells}, using 0 neighboring cells produced the highest F2 score (0.8481).

\begin{table}[ht]
  \centering
  \begin{minipage}[t]{0.48\textwidth}
    \vspace{0pt}  % Ensures top alignment
    \centering
    \begin{tabular}{p{0.6\linewidth}c}
      \toprule
      Number of neighbors & F2 Score \\
      \midrule
      0 & \textbf{0.8481} \\
      2 & 0.8316 \\
      4 & 0.8080 \\
      \bottomrule
    \end{tabular}
    \caption{Effect of using neighboring grid cells on detection performance.}
    \label{tab:neighboring-cells}
  \end{minipage}%
  \hfill
  \begin{minipage}[t]{0.48\textwidth}
    \vspace{0pt}  % Ensures top alignment
    \centering
    \begin{tabular}{p{0.6\linewidth}c}
      \toprule
      IoU loss & F2 Score \\
      \midrule
      CIoU & 0.8316 \\
      DIoU & 0.8321 \\
      IoU & 0.8293 \\
      GIoU & \textbf{0.8340} \\
      \bottomrule
    \end{tabular}
    \caption{Comparison of different IoU-based loss functions. We used 2 neighbors for this experiment.}
    \label{tab:iou-loss-comparison}
  \end{minipage}
\end{table}

Adding more neighboring cells led to a decrease in performance, suggesting that the decrease in precision is too high.

Next, we compared different IoU-based loss functions to determine their effectiveness in our model. Table \ref{tab:iou-loss-comparison} shows that the generalized IoU (GIoU) loss yielded the best performance with a F2 score of 0.8340.

However, the differences at F2 are quite small. This parameter therefore has no major influence on the result.

\subsection{Impact of Image Size and training techniques}

Table \ref{tab:image-size-comparison} evaluates the impact of varying image sizes on detection performance. Training on images of size 2048 provided the highest F2 score (0.8316).

\begin{table}[ht]
  \centering
  \begin{minipage}[t]{0.48\textwidth}
    \vspace{0pt}  % Ensures top alignment
    \centering
    \begin{tabular}{p{0.6\linewidth}c}
      \toprule
      Image size & F2 Score \\
      \midrule
      1024 & 0.7863 \\
      2048 & \textbf{0.8316} \\
      4096 & 0.8243 \\
      \bottomrule
    \end{tabular}
    \caption{Effect of different image sizes on detection performance. We used 2 neighbors and CIoU for this experiment.}
    \label{tab:image-size-comparison}
  \end{minipage}%
  \hfill
  \begin{minipage}[t]{0.48\textwidth}
    \vspace{0pt}  % Ensures top alignment
    \centering
    \begin{tabular}{p{0.7\linewidth}c}
      \toprule
      Method & F2 Score \\
      \midrule
      Baseline (Ours) & \textbf{0.848} \\
      + Mosaic Augmentation & 0.786 \\
      + Gradient Accumulation & 0.838 \\
      + No Maximum Size Constraint & 0.841 \\
      \bottomrule
    \end{tabular}
    \caption{Comparison of approaches to increase F2.}
    \label{tab:training-techniques}
  \end{minipage}
\end{table}

This confirms that we do not need the full resolution of 54000 x 31000 to find the vessel elements. Therefore, it is also not necessary to split the images to perform the detection for individual patches. Since only a single image needs to be predicted with our approach, we have a higher prediction speed.

We have successfully trained a model with a resolution of 6144 x 6144 on an A100 GPU with 40 GB VRAM. Even higher resolutions are possible with further adjustments to the architecture. It is important to emphasize that our standard model, which operates at a resolution of 2048 x 2048, is designed to be more accessible. It can be trained on consumer-grade hardware and requires only about 8 GB of VRAM for training.

In training our YOLO-based model, we explored several advanced techniques to enhance performance, including mosaic augmentation and gradient accumulation. Mosaic augmentation is a data augmentation strategy that creates a new training image by combining four different images from the dataset. This technique is intended to provide more context and variability during training, potentially improving the model's generalization ability. However, as shown in \cref{tab:training-techniques}, mosaic augmentation did not lead to an improvement in the F2 score for our task.

Gradient accumulation is another technique we evaluated. It allows for effective training with larger batch sizes than can fit in GPU memory by accumulating gradients over multiple mini-batches before updating the model weights. Despite its potential to stabilize training and improve convergence, our results indicate that gradient accumulation did not provide a significant benefit in our experiments.

One key modification that proved beneficial was the implementation of a maximum object width and height (the previously discussed anchor box variant). Removing this constraint resulted in a noticeable decrease in the F2 score, demonstrating the effectiveness of this technique in improving detection performance.

\subsection{Summary of the results}

We have demonstrated that WoodYOLO outperforms other YOLO variants in our specific use case. Interestingly, certain techniques that have consistently shown improvements in mAP on COCO do not yield similar benefits here. For instance, mosaic augmentation, introduced in YOLOv4 \citep{bochkovskiy2020yolov4optimalspeedaccuracy}, showed a 1.8\% increase in $\text{AP}_{50}$ in their ablation study. In contrast, our experiments reveal a substantial decrease of 6.2\% in F2 score when applying this technique. Similarly, we observed no advantage in using multi-positives, despite YOLOX reporting a 2.1\% improvement.

We attribute these discrepancies to several factors:

\begin{itemize}
    \item Metric difference: Our focus is on recall and approximate bounding box overlap, rather than the standard COCO metrics.
    \item Task simplification: As we only need to localize objects, our architecture can be shallower compared to those designed for more complex tasks.
    \item Reproducibility challenges: Deep learning, particularly in object detection, often faces reproducibility issues. Many YOLO implementations use legacy code with undocumented workarounds to improve AP, which are not mentioned in the original papers. These may include arbitrary loss function weightings or different weight decay strategies \citep{he2018bagtricksimageclassification}.
\end{itemize}

To mitigate these confounding factors, we developed our detector from scratch, avoiding reliance on previous codebases. This approach allows us to more accurately assess the impact of individual modifications.

In conclusion, our findings suggest that for specialized domains that diverge significantly from the standard COCO use-case, developing customized detectors can be more beneficial than adapting existing general-purpose models. This approach enables a more tailored solution that better addresses the specific requirements of the task at hand.

\section{Discussion and Conclusion}

In this paper, we presented WoodYOLO, a novel object detection algorithm specifically designed for microscopic wood fiber analysis. Our approach builds upon the YOLO architecture, incorporating tailored optimizations to enhance performance in high-resolution microscopy images. We introduced several key innovations, including a customized YOLO-based architecture optimized for microscopic images and a novel anchor box specification method.

Our comprehensive evaluation demonstrated that WoodYOLO outperforms state-of-the-art models such as YOLOv10 and YOLOv7 by significant margins in terms of F2 score. We also provided insights into the effectiveness of various architectural decisions and training techniques in the context of wood vessel detection.

The superior performance of WoodYOLO in detecting vessel elements in microscopic images of fibrous materials represents a significant advancement in automated wood species identification. This contribution has far-reaching implications for enhancing regulatory compliance, supporting sustainable forestry practices, and promoting biodiversity conservation efforts globally.

\subsection{Future Work}

The development of WoodYOLO opens up several promising avenues for future research and improvement. A key area for exploration is the integration of rotated bounding boxes to improve the accuracy of vessel element localization, particularly for elongated or angled structures. This further development requires adjustments to both the model architecture and the dataset annotations and offers considerable potential for improving detection accuracy.

At the same time, further optimization of the WoodYOLO architecture can be worked on to reduce the GPU requirements and increase the recall. Reducing the model's memory requirements is crucial to enable the processing of larger, higher-resolution microscopic images.

In addition to these technical improvements, we see great potential for adapting WoodYOLO to other areas that require high-resolution image analysis. For example, our approach could be useful in medical imaging to detect cell structures or in analyzing satellite imagery to identify specific geographical features. By exploring these cross-domain applications, we aim to extend the impact of our research beyond forestry and wood science and potentially contribute to advances in various scientific disciplines.

\bibliographystyle{unsrtnat}
\bibliography{references}

\end{document}